\documentclass{article}

\usepackage{arxiv}

\usepackage[utf8]{inputenc} 
\usepackage[T1]{fontenc}    
\usepackage{hyperref}       
\usepackage{url}            
\usepackage{booktabs}       
\usepackage{amsfonts}       
\usepackage{nicefrac}       
\usepackage{microtype}      
\usepackage{lipsum}
\usepackage{times}
\usepackage{epsfig}
\usepackage{graphicx}
\usepackage{amsmath}
\usepackage{amssymb}
\usepackage{siunitx}
\usepackage[font=footnotesize]{subfig}
\usepackage{float}
\usepackage{url}
\usepackage{tabularx}
\usepackage{gensymb}
\usepackage{mathtools}
\usepackage{blindtext}
\usepackage{enumitem}
\usepackage{amsfonts}
\usepackage[font=footnotesize]{subfig}

\title{Efficient 2D neuron boundary segmentation with local topological constraints}

\author{
  Thanuja D.~Ambegoda \\
  Institute of Neuroinformatics\\
  University of Zurich and ETH Zurich\\
  Switzerland \\
  \texttt{thanuja@ini.ethz.ch} 
   \And
 Matthew Cook \\
  Institute of Neuroinformatics\\
  University of Zurich and ETH Zurich\\
  Switzerland \\
  \texttt{cook@ini.ethz.ch}
}

\begin{document}
\maketitle

\begin{abstract}
We present a method for segmenting neuron membranes in 2D electron microscopy imagery. This segmentation task has been a bottleneck to reconstruction efforts of the brain's synaptic circuits. One common problem is the misclassification of blurry membrane fragments as cell interior, which leads to merging of two adjacent neuron sections into one via the blurry membrane region. Human annotators can easily avoid such errors by implicitly performing gap completion, taking into account the continuity of membranes.

Drawing inspiration from these human strategies, we formulate the segmentation task as an edge labeling problem on a graph with local topological constraints. 
We derive an integer linear program (ILP) that enforces membrane continuity, i.e.\ the absence of gaps. The cost function of the ILP is the pixel-wise deviation of the segmentation from a priori membrane probabilities derived from the data. 

Based on membrane probability maps obtained using random forest classifiers and convolutional neural networks, our method improves the neuron boundary segmentation accuracy compared to a variety of standard segmentation approaches. Our method successfully performs gap completion and leads to fewer topological errors. The method could potentially also be incorporated into other image segmentation pipelines with known topological constraints.
\end{abstract}

\section{Introduction}
\begin{figure}[!htbp]
\begin{center}
 	\centering
    \subfloat[]{\includegraphics[width=0.485\linewidth]{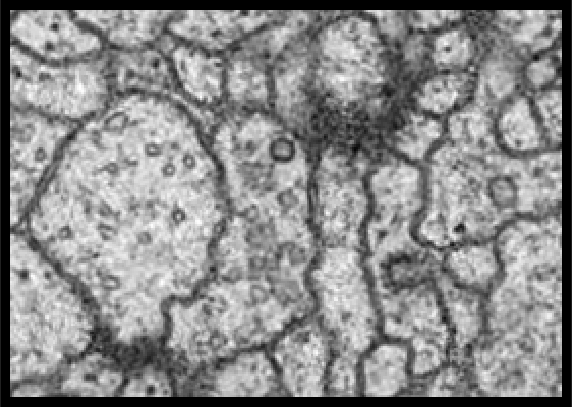}\label{subfig:raw_01}}
	\hfil
    \subfloat[]{\includegraphics[width=0.485\linewidth]{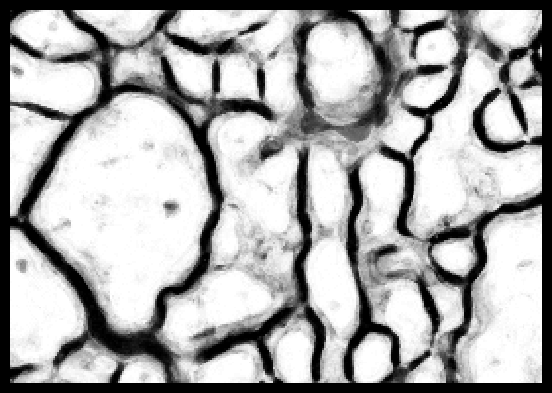}\label{subfig:membraneRFC}}
    \hfil	
    \vspace{-0.2cm}
    \subfloat[]{\includegraphics[width=0.485\linewidth]{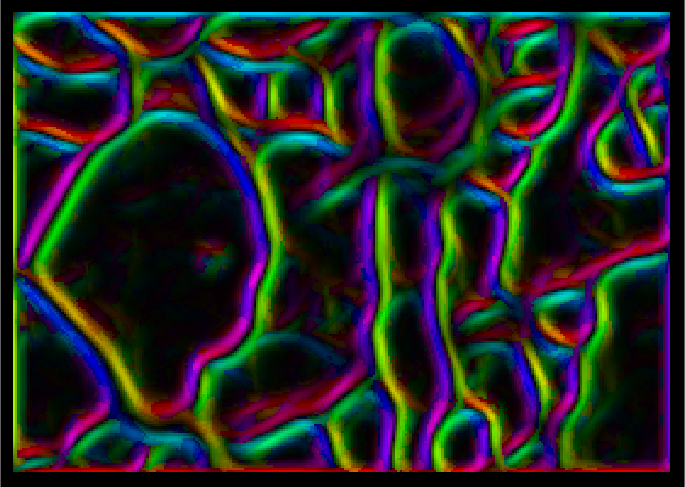}\label{subfig:ofr}}
    \hfil
    \subfloat[]{\includegraphics[width=0.485\linewidth]{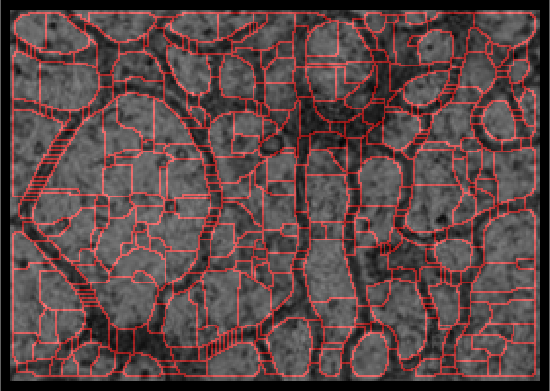}\label{subfig:ws}}
    \hfil
    \vspace{-0.2cm}
    \subfloat[]{\includegraphics[width=0.485\linewidth]{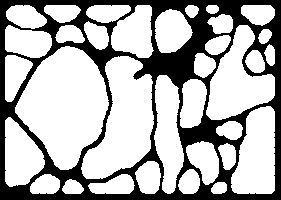}\label{subfig:membraneOut}}
    \hfil
    \subfloat[]{\includegraphics[width=0.485\linewidth]{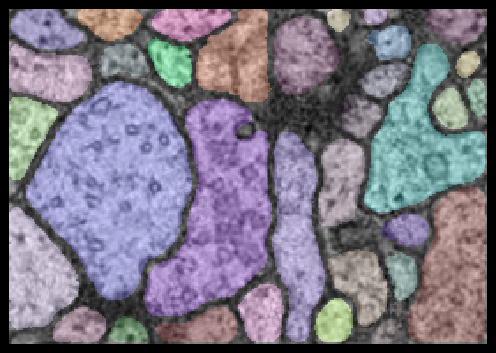}\label{subfig:segmentation}}

    \label{subfig:withconstr}
\end{center}
\vspace{-0.4cm}
\caption{\small Overview of our approach: 
\protect\subref{subfig:raw_01}~Raw EM image to be segmented.
\protect\subref{subfig:membraneRFC}~Neuron membrane probability map generated using a random forest classifier (white corresponds to low probability and black to high probability).
\protect\subref{subfig:ofr}~Max response from oriented edge filtering. The hue values $h \in [0,1]$ of each pixel encode the orientation $ \theta \in [0,360\degree]$ of the edge filter with the largest response.
\protect\subref{subfig:ws}~Graph of possible boundaries extracted using a watershed transform.
\protect\subref{subfig:membraneOut}~Membrane segmentation from our method. Note the gaps in the membranes in the probability map shown in \protect\subref{subfig:membraneRFC} are fixed in the segmentation.
\protect\subref{subfig:segmentation}~2D neuron slice labels obtained from our method.
}
\label{fig:overview}
\vspace{-0.4cm}
\end{figure}

Electron microscopy (EM) is a widely used technique for acquiring high-resolution imagery of neuronal tissue to visualize neuronal structures such as chemical synapses and vesicles (Fig.~\ref{subfig:raw_01}). To carry out research in neuro-anatomy, large quantities of EM images are acquired and then annotated manually~\cite{briggman2006, cardona2010}. Since manual annotation is a tedious process, there is a need for automatic and semi-automatic methods~\cite{fua2015, jain2010, kasthuri2015}. 

The most common problems in automatic neuron segmentation are caused by local ambiguities at neuron membranes when the algorithms erroneously merge two adjacent neurons into one. Such ambiguities are a result of low SNR in some parts of the image either because of imperfections in the image acquisition process or lack of sharpness of neuron membranes. Human annotators are much better at resolving such problems due to their ability of considering a larger context when dealing with local ambiguities. 
 
Therefore, one way for automatic segmentation methods to reach human level segmentation accuracy would be to take into account a larger context considering shape cues and continuation of structures in the presence of  gaps.
In this paper, we present a method to automatically annotate neuron slices (2D) on individual EM sections by accurately segmenting the neuron boundary that surrounds each neuron slice (Fig.~\ref{subfig:segmentation}).

Generic classifiers can be used to classify pixels of EM images into classes such as neuron membrane, synapses and mitochondria.
Pixelwise probability maps (Fig.~\ref{subfig:membraneRFC}) thus obtained are commonly used as the main input in many neuron segmentation approaches~\cite{andres2012,fua2015,funke2012,kasthuri2015,kaynig2015,gala2013,amelio2011}.There has been a significant improvement in pixel level image classification due to the recent progress in the area of deep neural networks, in particular with convolutional neural networks (CNN)~\cite{ciresan2012,tschopp2016}. 
The main drawback of CNNs is that they require a large training dataset which requires a significant amount of effort. When using CNNs, slight variations in the imaging parameters in datasets requires the CNNs to be trained separately for each dataset. 
Compared to CNNs, random forest classifiers (RFCs) require much less training data. However the quality of the probability maps generated by RFCs is usually less than those from CNNs. Our method produces significantly more accurate neuron segmentations using noisy probability maps from RFCs, compared to segmenting the same probability maps using standard techniques like the graph cut~\cite{boykov2001}, thereby providing a way of generating reasonably accurate segmentations using a small amount of training labels.

The automatic segmentation method we propose (Fig.~\ref{fig:overview}) uses topological constraints that reflect expected properties of an accurate segmentation of 2D neuron slices on an EM section. These constraints are defined on edges, nodes and faces of a planar graph which is derived from a membrane probability map. The segmentation problem is formulated as an integer linear program (ILP) that assigns an active or inactive state to each of the binary state variables (edges, nodes and faces of the graph) under these constraints. In addition to the constraints, we define an objective function that uses prior information about the state of these variables which is obtained by means of state of the art classifiers. Sec.~\ref{sec:graph} describes how we derive a graph of over-segmented object boundaries from a membrane probability map. Sec.~\ref{sec:ilp} to Sec.~\ref{sec:ilpobjective} presents the formulation of our segmentation method as an ILP. Sec.~\ref{sec:results} provides an evaluation of our approach with comparisons to other methods.

\section{Related work}

With the increase of automated and efficient EM image acquisition methods, there has been many advances in automated 3D neural circuit reconstruction methods in recent years~\cite{funke2012,amelio2011,kaynig2015}. Most of these methods commonly use pixelwise neuron membrane probabilities estimated using state of the art classifiers. 

The method presented in this paper focuses on segmenting neurons on 2D images which is potentially useful in 3D reconstruction pipelines where the accuracy of 3D reconstructions depend on the quality of 2D candidate segmentations~\cite{funke2012, funke2014}. Other methods that can be used to generate 2D segments include the graph cut and region merging approaches~\cite{gala2013,jain2011,andres2012}. Among the region merging approaches \cite{gala2013,andres2012} work in both 2D and 3D.

Our approach differs from the above methods because we use local constraints that focus on lowering topological errors which are usually caused by lower SNR, thereby addressing problems that are typically difficult for automatic methods. 

\section{Graphical representation of the segmentation problem}
\label{sec:graph}

\begin{figure*}[t!]
\centering
\hfill
\subfloat[]{\includegraphics[width=0.32\linewidth]{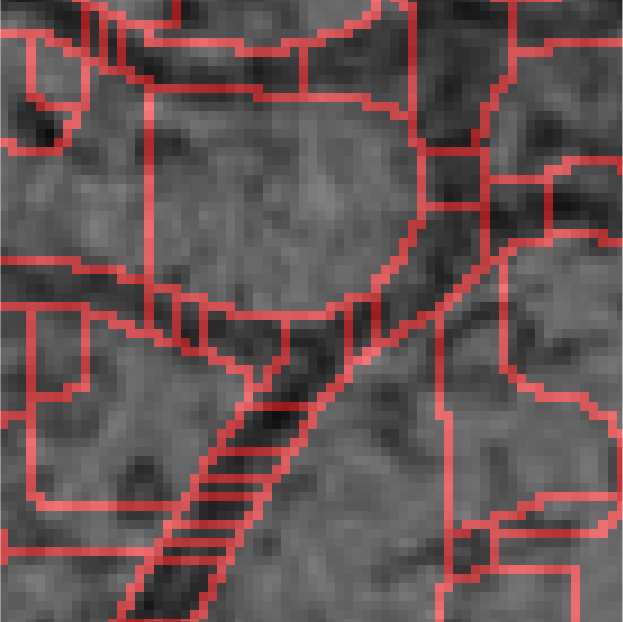}\label{subfig:zoomedWSgraph}}
\hfill
\subfloat[]{\includegraphics[width=0.32\linewidth]{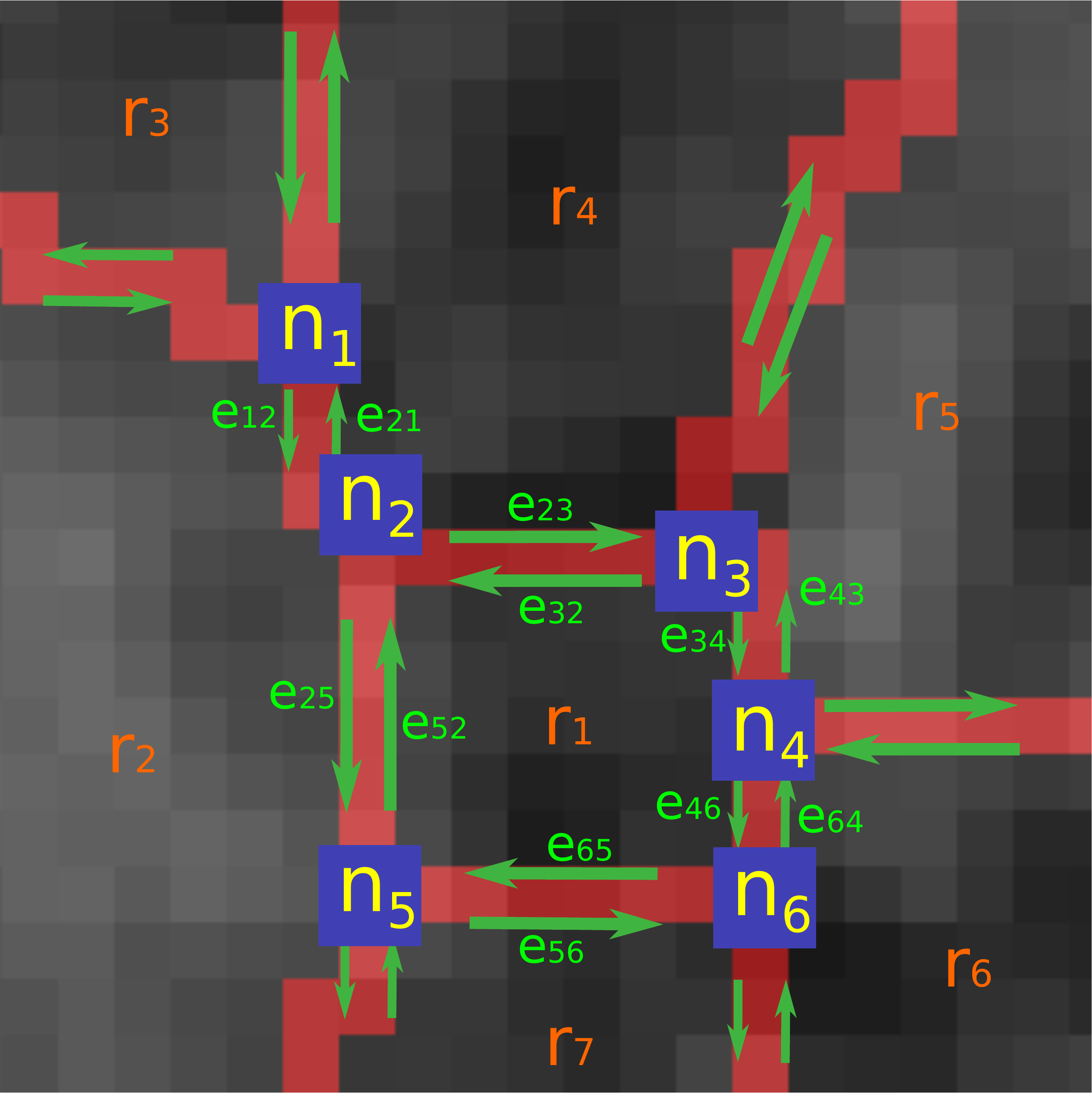}\label{subfig:graphSketch}}
\hfill
\subfloat[]{\includegraphics[width=0.32\linewidth]{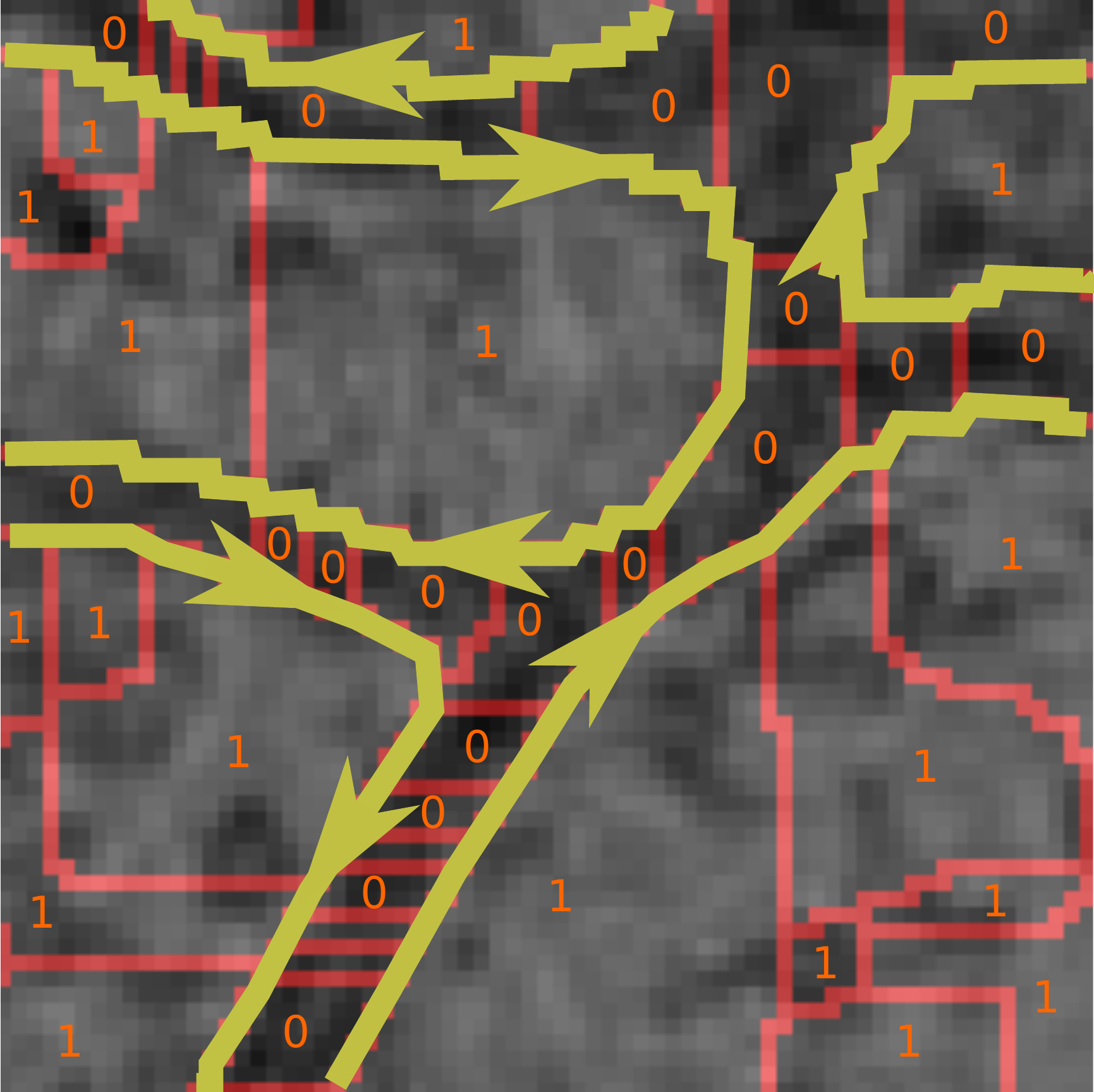}\label{subfig:ws_ilpState}}

\hfill
\caption{\small Illustration of the graph on which our optimization problem is defined: 
\protect\subref{subfig:zoomedWSgraph} Zoomed in version of Fig.~\ref{subfig:ws} illustrating the extracted planar graph representing possible object boundaries (red) that delineate cell interior and neuron membranes. This graph is overlaid on an EM image.
\protect\subref{subfig:graphSketch} Binary state variable types used in Sec.~\ref{subsec:variables}: \textit{edges (e), nodes (n)} and \textit{regions (r)} to define an integer linear program are annotated on a zoomed in version of \protect\subref{subfig:zoomedWSgraph}. Note that each object boundary in Fig.~\protect\subref{subfig:zoomedWSgraph} (shown in red) between any two nodes $n_i$ and $n_j$  is denoted by a pair of edges having opposite directions (green).
\protect\subref{subfig:ws_ilpState} Expected outcome of our segmentation task. Each region which is part of cell interior is assigned $1$ and membrane is assigned $0$. Each active (directed) edge separating cell interior and membrane is highlighted in yellow. All inactive edges remain red.
}
\label{fig:graphSketch}
\end{figure*}
Our method essentially represents the neuron segmentation task as an edge labeling problem on a planar graph (Fig.~\ref{subfig:ws}), where the edges of the graph correspond to potential object boundaries which delineate cell interior and membrane segments. The edge labeling procedure results in assigning a state to each of those edges. The assigned state would either be \textit{active} or \textit{inactive}, such that each active edge will be considered a true object boundary i.e.\ the two faces of the faces of the graph on either side of that edge will be assigned to opposite classes one of which is \textit{foreground} (cell interior) and the other \textit{background} (membrane).

Following sub-sections describe how we represent a membrane probability map of an EM image as a planar graph (Fig.~\ref{subfig:zoomedWSgraph}) which is then used to formulate an ILP. The optimal solution of the ILP states which edges (and regions) are active as shown in Fig.~\ref{subfig:ws_ilpState}.   

\subsection{Probability maps}
As the first step, we obtain a membrane probability map of neuron membranes corresponding to the input EM image (Fig.~\ref{subfig:membraneRFC}).
Each pixel of such a probability map gives the probability of that pixel belonging to the class of neuron membrane. These probability maps can be generated using any generic classifier of choice (e.g.\ CNN, RFC). 

\subsection{Graph of over-segmented object boundaries}
\label{subsec:graph}
In the second step potential object boundaries are extracted from probability maps. We use a bank of oriented edge detection filters (Fig.~\ref{subfig:filterElement}) to detect possible object boundaries along with their orientation. The output of such a filter bank applied to a membrane probability map quantifies how likely it is for an edge to be present at each pixel for each of the orientations defined in the edge filter bank. The filters have orientations ranging from $0 \degree $ to $350 \degree $ in steps of $10 \degree $. 
The maximal response from these filters at each pixel is illustrated in Fig.~\ref{subfig:ofr}. 

A watershed transform is applied on the filter output to obtain a height map of the local maxima of the max response of the filter. The ridges of the watershed transform corresponds to a graph of containing all possible edges detected by the oriented filter bank (Fig.~\ref{subfig:ws}). Therefore, this graph can be considered as an over-segmentation of object boundaries that we would proceed to segment. To avoid having too many small watershed super-pixels (a.k.a.\ fragments), the filter response is smoothened using a Gaussian filter with $\sigma = 1.4$ pixels, before applying the watershed transform.

\section{Problem representation using binary state variables}
\label{sec:ilp}
Using the graphical representation in Fig.~\ref{subfig:zoomedWSgraph} we formulate an optimization task that results in the desired labeling of edges and regions where the regions which are part of cell interior are assigned to foreground and the regions which are part of membrane are assigned to background as illustrated in Fig.~\ref{subfig:ws_ilpState}. 
This section describes how we formulate an ILP using the graph of over-segmented object boundaries such that the optimal solution corresponds to an accurate segmentation of neuron boundaries.

\label{subsec:variables}
We define three types of binary state variables corresponding to the edges, nodes and faces of that graph as shown in Fig.~\ref{subfig:graphSketch}.
The \textit{active state} of a binary state variable means that it has the value one.\\

\textbf{(i) Edge state binary variables}
Each edge shown in the graph (Fig.~\ref{subfig:zoomedWSgraph}) between two nodes $n_i$ and $n_j$ colored in red has three corresponding state variables: (a) $e_{ij} = 1$: directed edge exists from node $n_i$ to $n_j$, (b) $e_{ji} = 1$: directed edge exists in the opposite direction, (c) $e^0_{ij} = 1 \iff e_{ji}^0 = 1$: lack of an edge between nodes $n_i$ and $n_j$. 
If the edge between the nodes $i$d $j$ is part of the object boundary, either $e_{ij}$ or $e_{ji}$ is set to be active by the ILP (Fig.~\ref{subfig:ws_ilpState}).

\vspace{2mm}

\textbf{(ii) Region state binary variables}
Each face (region) $m$ of the graph has two corresponding state variables $r_m$ and $r_m^0$. $r_m = 1$ when region $m$ is part of foreground in the segmentation output. $r_m^0$ is active when region $m$ is part of background.

\vspace{2mm}

\textbf{(iii) Node state binary variables}
In an accurate segmentation output, the angles between two edges (neuron boundaries) tend to be smooth. Therefore, sharp angles occurring between two active edges are penalized in the ILP objective. Assigning such a penalty to the activation of a pair of edges gives rise to a quadratic program. In order to keep the problem formulation linear we define \textit{node state variables} which correspond to pairs of directed edges.
At any node of the graph (Fig.~\ref{subfig:zoomedWSgraph}) there are at least 3 edges (6 directed edges as in Fig.~\ref{subfig:graphSketch}) out of which exactly two or zero has to be active as described later in Sec.~\ref{subsec:ILPconstraints}. 

A set of node state variables $n_i^c$ are defined for each node. When a particular node state $n_i^c$ is active, it specifies which pair of directed edges are active out of all edges connected to this node. 

$n_i^{c=0}$ is the state where the node is not active i.e.\ not part of the segmentation and therefore none of the edges connected to it are active. Fig.~\ref{fig:nodeConfigs} illustrates the set of node state variables for a typical node of the graph with 3 neighbors.

\section{ILP constraints}
\label{subsec:ILPconstraints}
The constraints of an ILP are a set of bounds for different linear combinations of the state variables that define the solution space of the entire set of state variables.

We define a set of hard constraints that reflect general properties expected in an accurate segmentation output. Following is a description of these linear constraints used in our ILP.

\subsection{Low-level constraints}
We define an activation constraint for each variable type as follows.

\vspace{2mm}

\textbf{(i) Directed edge activation}
Between two adjacent nodes $n_i$ and $n_j$, we allow either the outgoing edge $e_{ij}$ (w.r.t $n_i$) to be active or the incoming edge $e_{ji}$ to be active or none of them to be active ($e_{ij}^0 = 1 \iff e_{ji}^0 = 1$).
Therefore, each edge of the graph in Fig.~\ref{subfig:zoomedWSgraph} has three corresponding binary edge states, out of which exactly one has to be set to $1$. The mathematical formulation of this constraint is as follows: 	

			\begin{equation}
			\forall e_{ij} \in \mathcal{E}, \; e_{ij} + e_{ji} + e_{ij}^0 = 1 ,
			\end{equation}
			where $\mathcal{E}$ is the set of all edges in the graph.
			
\vspace{2mm}

\textbf{(ii) Node activation}
\begin{figure}[t]
\begin{center}
    \centering
    \subfloat[]{\includegraphics[width=0.17\linewidth]{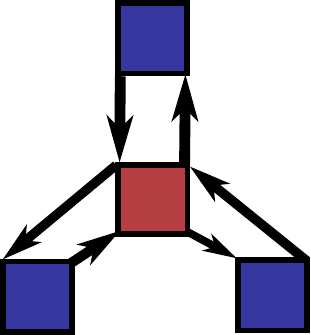}\label{subfig:node}}
    \hspace{1cm}
    \subfloat[]{\includegraphics[width=0.17\linewidth]{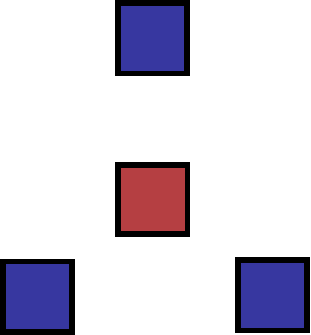}\label{subfig:node_0}}
    \hspace{1cm}
    \subfloat[]{\includegraphics[width=0.17\linewidth]{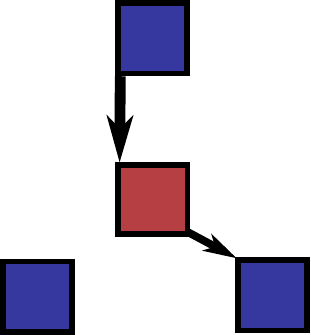}\label{subfig:node_1}}
    \hspace{1cm}
    \subfloat[]{\includegraphics[width=0.17\linewidth]{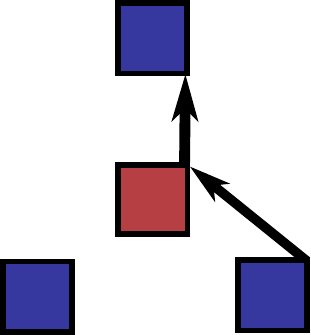}\label{subfig:node_2}}
    \hspace{1cm}
    \subfloat[]{\includegraphics[width=0.17\linewidth]{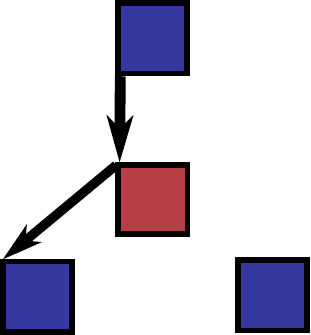}\label{subfig:node_3}}
    \hspace{1cm}
    \subfloat[]{\includegraphics[width=0.17\linewidth]{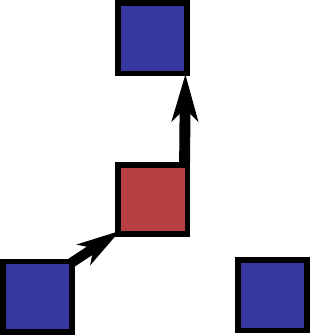}\label{subfig:node_4}}
    \hspace{1cm}
    \subfloat[]{\includegraphics[width=0.17\linewidth]{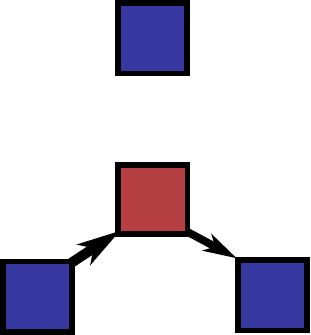}\label{subfig:node_5}}
    \hspace{1cm}
    \subfloat[]{\includegraphics[width=0.17\linewidth]{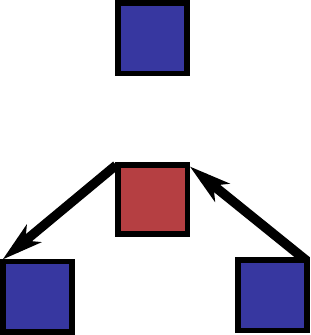}\label{subfig:node_6}}	
\end{center}
\vspace{-0.4cm}
\caption{\small \textit{Node states}: \protect\subref{subfig:node} depicts a typical node (in red) in a graph similar to Fig.~\ref{subfig:graphSketch} that is connected to each of its three neighboring nodes (blue) via a pair of directed edges. Fig.\protect\subref{subfig:node_0} to \protect\subref{subfig:node_6} are all possible $p$ node states for the node in red. Each of these states are represented by a binary variable in the ILP formulation. \protect\subref{subfig:node_0} shows the inactive node state where it is not connected to any of its neighbors. \protect\subref{subfig:node_1} to \protect\subref{subfig:node_6} show all possible active states this node is allowed to have according to the given constraints. In each of these active node states there are exactly two active edges attached to it, where one is incoming and the other out going. Therefore, $p=2 \cdot \binom{L}{2} + 1$ (where $L$ is the number of directly connected nodes to the current node). }
\label{fig:nodeConfigs}
\vspace{-0.4cm}
\end{figure}
Similarly, for each node $n_i$ we enforce that one of all possible nodes states ($c \in \{0,1,...,p\}$) must be active.
			\begin{equation}
			\forall n_i \in \mathcal{N}, \; 
            \sum_{c=0}^p n_{i}^{c} = 1,
			\end{equation}
			where $\mathcal{N}$ is the set of all nodes and $p$ is the number of binary node states of node $n_i$ as illustrated in Fig.~\ref{fig:nodeConfigs}.
			
\vspace{2mm}

\textbf{(iii) Region activation}
        Each face of the graph (region) is assigned two binary indicator variables: $r_m$ and $r_m^0$. When region $m$ is assigned to foreground, the corresponding region binary state variables should have the values: $r_m = 1$ and $r_m^0 = 0$.  		
Therefore, region state variables are activated according to the following constraint:		
		\begin{equation}
		\forall m \in \mathcal{R}, \; r_m + r_m^0 = 1,
		\end{equation}
where $\mathcal{R}$ is the set of all regions.	 
   
\subsection{Topological constraints} 
Co-activation of different combinations of state variables results in different topologies in the output segmentation. 

If not properly constrained, the resulting topology could deviate from what's expected in an accurate segmentation.
Therefore, we define the following constraints in order to ensure that resulting segmentation have an acceptable topology. 

\vspace{2mm}

\textbf{(i) Closed loop of edges}
\label{subsec:closedloopofedges}
\begin{figure}[t]
 	\centering
    \subfloat[]{\includegraphics[width=0.44\linewidth]{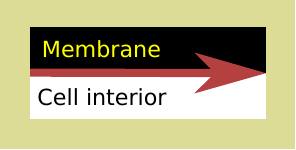}\label{subfig:filterElement}}
	\hfil
    \subfloat[]{\includegraphics[width=0.44\linewidth]{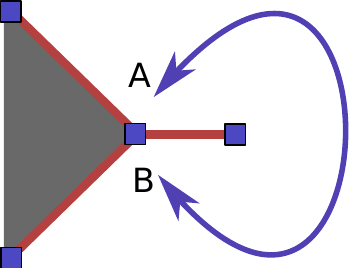}\label{subfig:danglingEdge}}   
\hfil
    \subfloat[]{\includegraphics[width=0.5\linewidth]{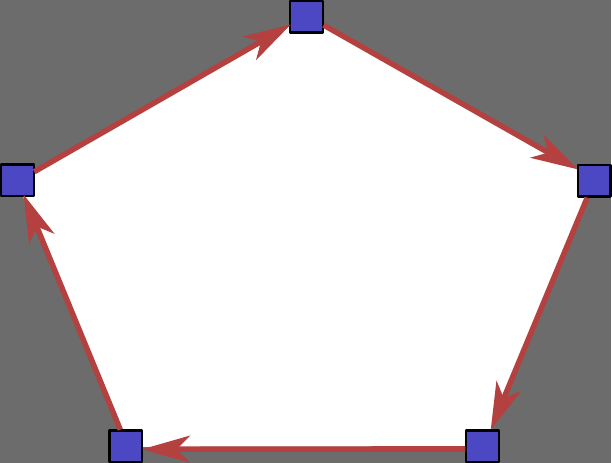}\label{subfig:clockwiseCycle}}
\caption{\small 
\protect\subref{subfig:filterElement}~Illustration of the oriented edge detection filter element used in Sec.~\ref{subsec:graph}. Cell interior (foreground) is always on the right hand side of the directed edge.
\protect\subref{subfig:danglingEdge}~A dangling edge would mean that the regions A and B on either side of the edge are of the same type. This is a topological error because an edge has to have two opposite types of regions on its sides (foreground and background). We prevent this using the constraint \textit{closed loop of edges} (Eq.~\ref{constraint:closedLoopOfEdges}). 
\protect\subref{subfig:clockwiseCycle}~Because of our definition of a directed edge, any foreground segment has to be bounded by a closed clockwise loop of active edges. We enforce this topological property using constraint \textit{clockwise edge loop around foreground} (Eq.~\ref{constraint:clockwiseEdgeLoop}).
}
\label{fig:graph_elements}
\end{figure}
In the segmentation output each foreground segment must be enclosed by a closed loop of edges.
This also means there must not be any dangling edges activated, that do not have exactly two other edges connected to either end of it (Fig.~\ref{subfig:danglingEdge}). 
We enforce this topological property as a constraint as follows:
\begin{itemize}[noitemsep]
\item When a node is active exactly two edges connected to it have to be active as well.
\item When the node $n_i$ is inactive ($n_{i}^{0} = 1$), none of the edges attached to it can be active.
\end{itemize}
This results in the following mathematical formulation:
			\begin{align}
			\forall n_i \in \mathcal{N}, \; 2 n_{i}^{0} +  \smashoperator{\sum_{e_{ij} \in {E_{n_i}}} }e_{ij} = 2,   
			\label{constraint:closedLoopOfEdges}
			\end{align}
		where $E_{n_i}$ is the set of edges attached to node $n_i$. Any neighbor of $n_i$ connected via the edges $e_{ij}$ and $e_{ji}$ is denoted by $n_j$.	

\vspace{2mm}

\textbf{(ii) Activate corresponding edges of a node state} 
For each active state of node $n_{i}$, there is a unique pair of directed edges that have to be activated. While constraint \eqref{constraint:closedLoopOfEdges} ensures there are exactly two active edges for an active node, this constraint (Eq.~\eqref{constraint:nodestate_edgestate_correspondence}) ensures the matching between the predefined node states and the corresponding active edge pairs. This matching is important because in the objective function (Sec.~\ref{sec:ilpobjective}) we use a precalculated set of costs for each node state based on the angle between its active pair of edges.  

Consider a typical node $n_i$ with (at least) a pair of neighboring nodes $n_j$ and $n_k$. Let node state $n_i^c$ have the edges $e_{ij}$ (outgoing edge) and $e_{ki}$ (incoming edge) activated.
		
        We formalize this constraint as follows:
			\begin{equation}
			\forall n_i \in \mathcal{N} , \forall c \in \mathcal{C}_{n_i}, \; 0 \leq -2 n_{i}^{c} + e_{ij} + e_{ki} \leq 1,
			\label{constraint:nodestate_edgestate_correspondence}
			\end{equation}
            where $\mathcal{N}$ is the set of all nodes and $\mathcal{C}_{n_i}$ is the set of all node states of node $n_i$.

\vspace{2mm}

\textbf{(iii) Clockwise edge loop around foreground}
As mentioned previously, topological constraint (i) given by Eq.~\eqref{constraint:closedLoopOfEdges} enforces a closed cycle of edges around every foreground segment. 
However according to the definition of a directed edge (Fig.~\ref{subfig:filterElement}) foreground must occur to the right of an active edge. Therefore, these closed cycle of edges should have a clockwise sense (Fig.~\ref{subfig:clockwiseCycle}). 

This topological constraint is formulated in the ILP as follows:
			\begin{equation}
			\forall e_{ij} \in \mathcal{E}, \; - e_{ij} + e_{ji} + r_{j}^1 - r_{i}^1 = 0.
            \label{constraint:clockwiseEdgeLoop}		                 
			\end{equation}	

\vspace{2mm}

\textbf{(iv) Membranes as closed loops}
\label{subsec:membraneContinuityConstraint}
We do not expect to see gaps in the segmented neuron membranes. In other words, as shown in Fig.~\ref{subfig:ws_ilpState}, the segments which are assigned to be membrane (background) are contiguous. Any region which is assigned to background must have at least two other adjacent regions connected to it that are also assigned to background.  This property is formalized as a constraint using the following equation:
\begin{equation}
	\forall r_i \in \mathcal{R}, \; r_{i}^1 - r_{i}^0 + \sum_{e_j \in E_{r_i}} e^0_j \geq 1.	
    \label{constraint:membraneContinuity}
\end{equation}				
		Here, $e_j$ is an element of the set $E_{r_i}$ of all edges bounding region $r_i$ and $e_j^0 = 1$ when edge $e_j$ is not active. In Eq.~\eqref{constraint:membraneContinuity} making sure that at least 2 of the edges are inactive when the region is turned off (i.e. assigned to background), we enforce that region to be part of a contiguous membrane segment. 

We observe that this constraint helps the system to fill gaps observed in membrane probability maps (Fig.~\ref{subfig:membraneRFC}). This is useful since neuron membranes sometimes gets blurred out and leads to merging of two adjacent neuron sections via this blurry region due to that region being misclassified as being part of cell interior.  

\section{ILP objective function}
\label{sec:ilpobjective}
The purpose of the objective function of our ILP is to minimize the deviation of the optimal solution from prior information derived from the data. We precompute coefficients for each binary state variable using the input data, using classifiers that are trained on available ground truth.
		The objective function of the ILP is:
		\begin{equation} \label{eq:ILPobjective}
		\{ e^*,n^*,r^* \} = \operatornamewithlimits{argmin}\limits_{e \subset \mathcal{E},n \subset \mathcal{N},r \subset \mathcal{R}} \: C_{e} + C_{n} + C_{r}.
		\end{equation}
		
\begin{figure}[t]
 	\centering
    \subfloat[]{\includegraphics[width=0.485\linewidth]{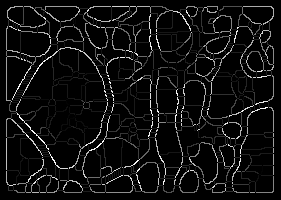}\label{subfig:edgeScores}}
	\hfil
    \subfloat[]{\includegraphics[width=0.485\linewidth]{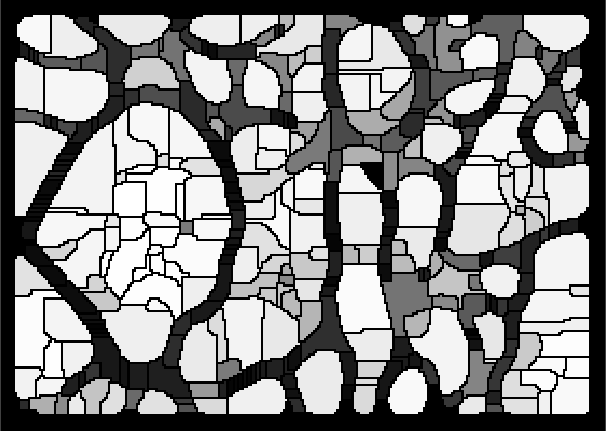}\label{subfig:regionScores}}
\caption{\small Visualization of \textit{a priori} probabilities associated with \textit{edge state variables} and \textit{region state variables} of an EM image, which are used in the ILP objective (Equation.\ \ref{eq:ILPobjective}).
\protect\subref{subfig:edgeScores} Each pixel belonging to an edge indicates the \textit{a priori} probability of that edge being at the boundary between neuron membrane and cell interior. These probabilities are generated using a random forest classifier. White indicates a high probability.
\protect\subref{subfig:regionScores} Each region is assigned an \textit{a priori} probability that it belongs to cell interior. This probability is obtained from the membrane probability map by averaging the pixelwise probabilities over each region.}
\label{fig:unaryScores}

\end{figure}

		Here, the terms $C_{e}$, $C_{n}$ and $C_{r}$ represent the weighted cost terms for the activation of the state variables corresponding to edges, nodes and regions in the model respectively. The objective term $C_{e}$ for edge state variables is defined as:
		\begin{equation} \label{eq:edgecosts}		
		C_{e} = \sum_{e_i \in \mathcal{E}} w_{e}^{on} \cdot u_{e_{ij}} ( e_{ij} + e_{ji} ) + w_{e}^{off} \cdot ( 1-u_{e_{ij}} ) \cdot e^0_{ij}, 		
		\end{equation}
		
		where $w_{e_{on}}$ and $w_{e_{off}}$ are the linear weights controlling  the relative reward (or penalization) of turning an edge on (or off). $u_{e_{ij}}$ is the \textit{a priori} probability for edge $e_{ij}$ to be active i.e.\ be part of an object boundary. This probability can be learned using any general classifier of choice. Fig.~\ref{subfig:edgeScores} illustrates \textit{a priori} edge activation probabilities obtained using a random forest classifier. 
        
		The cost term $C_{n}$ corresponding to all node state variables $n_i \in \mathcal{N}$ is:
		\begin{equation} \label{eq:nodeCostTerm}
		C_{n} = \sum_{n_i \in \mathcal{N}} \{ w_{n}^{off} \cdot u_{n_i^{0}} \cdot n_i^0 + w_{n}^{on} \sum_{c \in \mathcal{C}_i } u_{n_{i}^{c}} \cdot n_i^c  \},
		\end{equation}
        where $\mathcal{C}_i$ is the set of all node states of node $n_i$.
        
	\begin{figure}[t]
  	\centering
	\includegraphics[width=5.0cm]{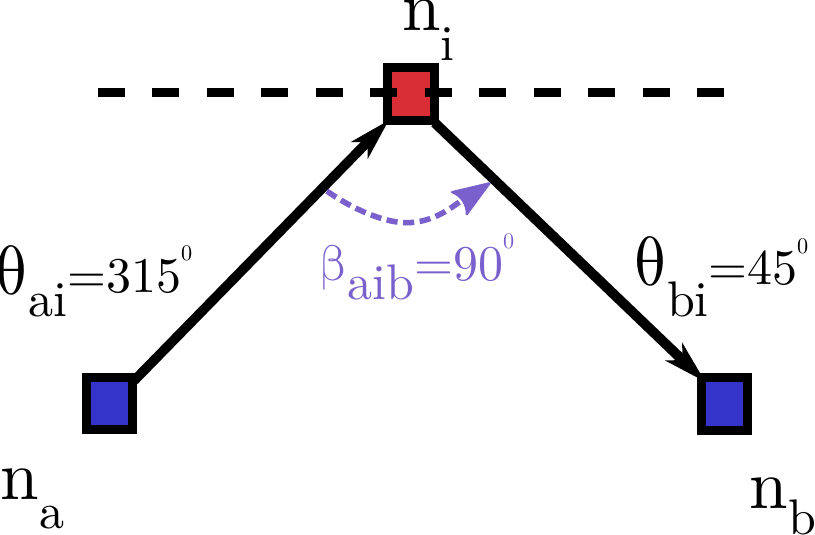}
    \caption{\small Illustration of node angle $\beta_{aib}$ used in Equation \ref{eq:nodeCost}. $\theta$ values are obtained from \textit{oriented edge filters} as described in Sec.~\ref{subsec:graph}. The horizontal dotted line is the reference with respect to which the angles are defined for the oriented edge filters.}
    \label{fig:nodeAnglesNotation}
    \end{figure}
		In Eq.~\eqref{eq:nodeCostTerm} we have set the score $u_{n_i^{0}}$ for any inactive node configuration to be equal to $1$ so that in the ILP objective its contribution is solely determined by the weight $w_n^{off}$. The score $u_{n_i^{c}}$ assigned to each active node configuration $c$ of node $n_i$ is calculated as

\begin{equation}
\label{eq:nodeCost}
u_{n_i^{c}} = f(\beta_{aib}) = \frac{1}{\sigma \sqrt{2 \pi} } 
			\: e^{ - \frac{(\beta - \pi)^2 }{2 \sigma^2} }
\end{equation}

where $\beta_{aib}$ is the angle between the two active edges corresponding to the active node configuration $j$ of node $n_i$ (Fig.~\ref{fig:nodeAnglesNotation}). This score is a smoothness term which is maximal for a node configuration where the incoming and outgoing edges are at an angle $180 \degree$ and drops to zero when it deviates from $180 \degree$. The variation of the smoothness is modeled using a Gaussian function with mean $180 \degree$ and a predefined spread $\sigma$ ($\approx 45 \degree$) as given in equation \ref{eq:nodeCost}.
       
The objective term for all regions $C_{r}$ is:
		
		\begin{equation} \label{eq:regionCostTerm}
			C_{r} = \sum_{r_i \in \mathcal{R}}  w_{r_i}^{on} \cdot u_{r_i} \cdot  r_i + \sum_{r_i^0 \in \mathcal{R}} w_{r_i}^{off} \cdot u_{r_i^0} \cdot r_i^0
		\end{equation}
where $u_{r_i}$ is the \textit{a priori} probability of the region being part of cell interior. This value is obtained by averaging the membrane probability of all the pixels in a given region (Fig.~\ref{subfig:regionScores}).

We use the structured learning framework suggested in \cite{smola2007,teo2010} to learn the optimal parameters $w_e^{on}$, $w_e^{off}$, $w_n^{on}$, $w_n^{off}$, $w_r^{on}$ and $w_r^{off}$ of the linear objective function. Just one image of size $512 \times 512$ pixels with ground truth labeling was sufficient for parameter learning.     
\section{Results}
\label{sec:results}
\begin{figure}[t]
\centering
\includegraphics[width=\linewidth]{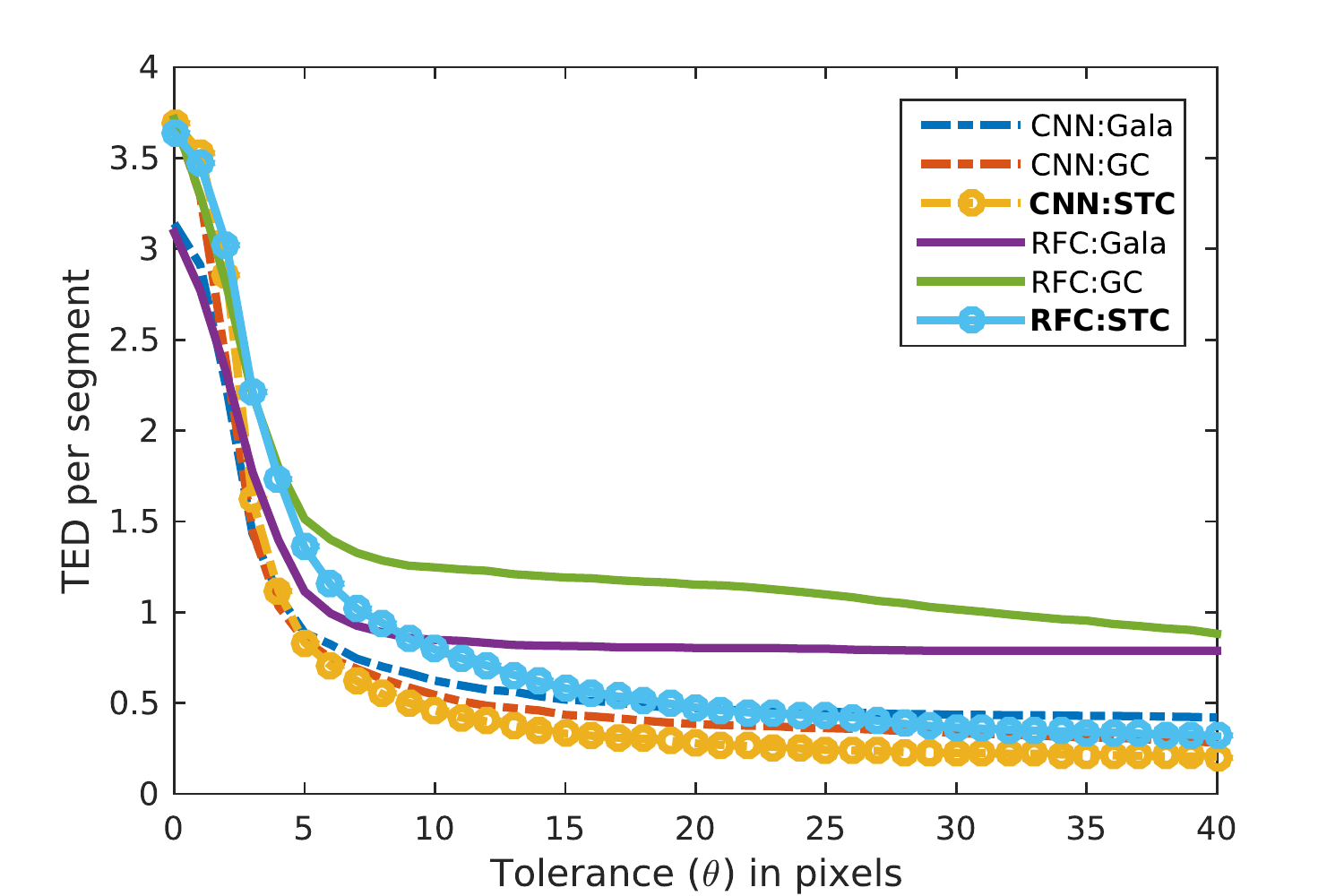}
\caption{\small Total topological error quantified using TED plotted against the boundary tolerance allowed in pixels.
which is the sum of false positives, false negatives, false splits and false merges. normalized by the number of segments in ground truth. Segmentation using our approach with CNN probability maps as inputs (CNN:STC) shows the best accuracy.
}
\label{fig:tedAll}
\end{figure}
\begin{figure}[t]
 	\centering
    \subfloat[]{\includegraphics[width=0.485\linewidth]{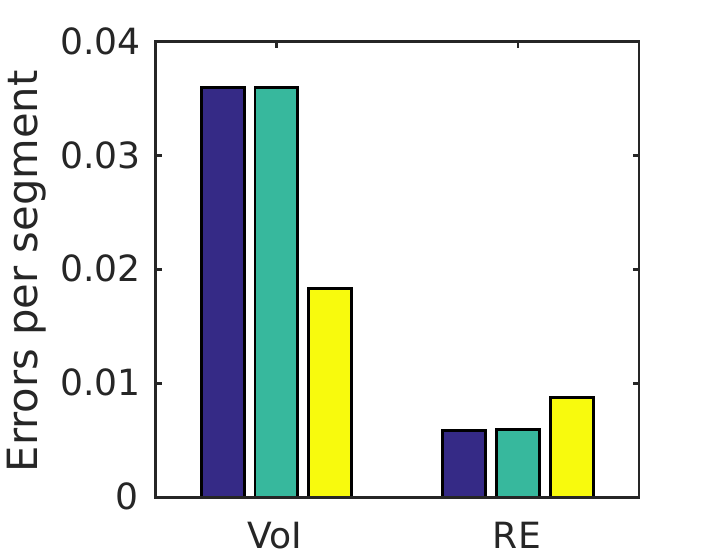}\label{subfig:rfc_re_voi}}
	\hfil
    \subfloat[]{\includegraphics[width=0.485\linewidth]{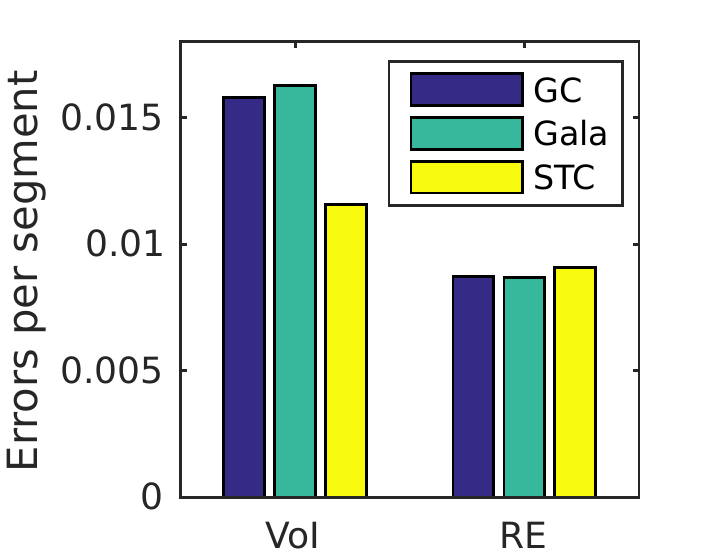}\label{subfig:cnn_re_voi}}

\caption{\small Comparison of segmentation errors using Variation of Information (VoI) and Rand Error ($RE = 1- RI$).
Methods compared: Graphcut (GC)~\cite{boykov2001}, Graph-based active learning of agglomeration (GALA)~\cite{gala2013} and our method (STC).
\protect\subref{subfig:rfc_re_voi} Using RFC probability maps as inputs.
\protect\subref{subfig:cnn_re_voi} Using CNN probability maps as inputs.}
\label{fig:re_voi_ted}
\end{figure}
\textbf{Evaluation criteria}

We use Rand Index (RI)~\cite{rand1971}, Variation of Information (VoI)~\cite{meila2005} and Tolerant Edit Distance (TED)~\cite{funke2016} to evaluate our approach against other image segmentation methods which are also based on clustering pixels using probability maps.
Both RI and VoI are known to be affected by boundary shifts of the segmentations, even though such deviations will not affect the topological form of the segmentation that we are interested in \cite{jain2010,funke2016}. TED tries to avoid this problem by counting the minimum topological errors within a specified boundary shift of segments. The topological errors evaluated are: false splits, false merges, false positives (foreground segment identified where there should be none), false negatives (missed segment).

\textbf{Experiments}
\begin{figure}[t]
 	\centering
    \subfloat[]{\includegraphics[width=0.485\linewidth]{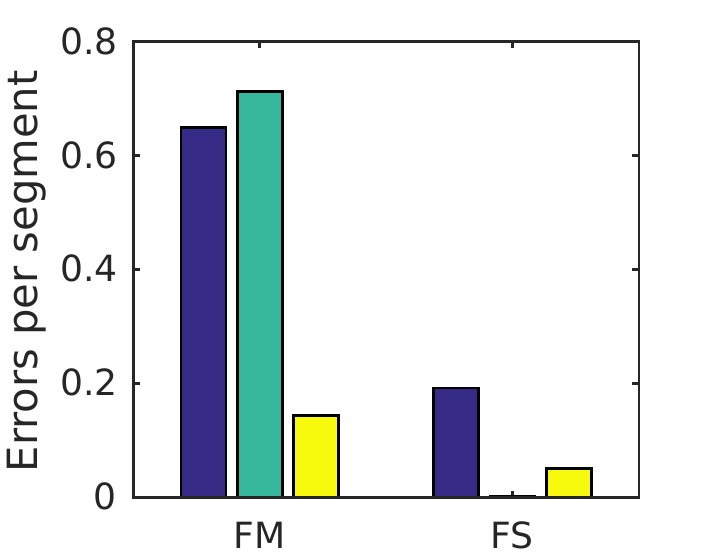}\label{subfig:rfc_fmfs}}
	\hfil
    \subfloat[]{\includegraphics[width=0.485\linewidth]{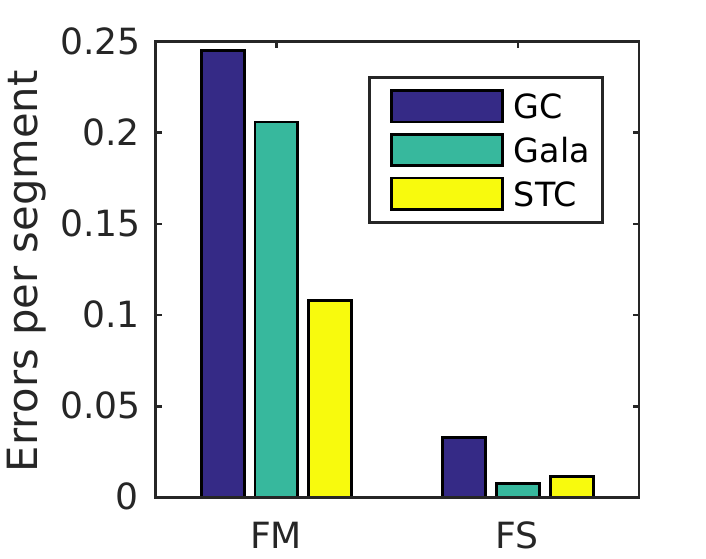}\label{subfig:cnn_fmfs}}
\caption{\small Breakdown of topological segmentation errors quantified using Tolerant Edit Distance (TED)~\cite{funke2016} comparing Graphcut (GC)~\cite{boykov2001}, Graph-based active learning of agglomeration (GALA)~\cite{gala2013} and our method (STC). 
\protect\subref{subfig:rfc_fmfs} False merges and false splits using RFC inputs.
\protect\subref{subfig:cnn_fmfs} False merges and false splits using CNN inputs.}
\label{fig:ted_errors}
\end{figure}
We evaluate our method using a publicly available dataset\cite{cardona2010} of neural tissue of Drosophila larva, along with manually annotated neuron membrane labels. This dataset contains 30 sections of serial section Transmission Electron Microscopy (ssTEM) images of size $512 \times 512$ pixels. Each pixel is of dimensions $\SI{4}{nm}  \times \SI{4}{nm}$. Each section is approximately $\SI{50}{nm}$ thick. We set aside images $21$ to $30$ for evaluations.

\textbf{Probability maps} We used the random forest classifier provided by \textit{Ilastik} segmentation toolkit~\cite{ilastik2011}. The classifier was trained interactively by placing a few brush strokes to mark neuron membrane and cell interior separately on the first $3$ images of the dataset. Another set of probability maps were generated using the CNN implementation provided by \cite{tschopp2016}, using the first $20$ images of the same dataset for training. 

\textbf{Comparisons}
We compare the performance of our method with the graphcut (GC)~\cite{boykov2001} and Graph-based active learning of agglomeration (GALA)~\cite{gala2013} using probability maps generated as mentioned above. We have a lower VoI than both GC and GALA. Furthermore, our approach results in a lower cumulative topological error than GC and GALA (Fig.~\ref{fig:tedAll}).

\section{Discussion}
We have proposed a method to segment neuron membranes in EM images which results in fewer topological errors. Our method takes pixelwise membrane probability maps as inputs which are then used to represent the segmentation task as an edge labeling problem on a graph. The edge labeling is solved by formulating an ILP using topological constraints to improve segmentation accuracy. 

We have shown that our method can be used to produce 2D neuron segmentations using probability maps from both CNNs and RFCs with less topological errors than other segmentation methods.

State of the art classifiers based on deep learning such as CNNs that produce high quality probability maps for EM datasets need a lot of training labels for each dataset, which is a major bottleneck for automated segmentation methods. On the other hand, random forest classifiers require much less training data with the drawback of resulting in probability maps of lower quality.  

Due to the generic nature of the topological constraints used in our approach, we note that this method can be potentially used in any image segmentation pipeline that have similar properties.

\section*{Acknowledgment}
This research was funded by the National Competence Center for Biomedical Imaging (NCCBI) of Switzerland. The authors would like to thank Julien Martel, Jan Funke and Richard Hahnloser for their ideas and feedback.

\bibliographystyle{unsrt}  
\bibliography{references}  

\begin{thebibliography}{10}

\bibitem{briggman2006}
K.~L. Briggman and W.~Denk.
\newblock Towards neural circuit reconstruction with volume electron microscopy
  techniques.
\newblock {\em Current Opinion in Neurobiology}, 16(5):562--570, 2006.

\bibitem{cardona2010}
A.~Cardona, S.~Saalfeld, S.~Preibisch, B.~Schmid, A.~Chen, J.~Pulokas,
  P.~Tomancak, and V.~Hartenstein.
\newblock An integrated micro- and macroarchitectural analysis of the
  drosophila brain by computer-assisted serial section electron microscopy.
\newblock {\em PLoS Biology}, 8(10), 2010.

\bibitem{fua2015}
P.~Fua and G.~Knott.
\newblock Modeling brain circuitry over a wide range of scales.
\newblock {\em Frontiers in Neuroanatomy}, 2015.

\bibitem{jain2010}
V.~Jain, H.~S. Seung, , and S.~C. Turaga.
\newblock Machines that learn to segment images: a crucial technology for
  connectomics.
\newblock {\em Curr. Opin Neurobiol.}, 20:653--66, 2010.

\bibitem{kasthuri2015}
N.~Kasthuri, K.~J. Hayworth, D.~R. Berger, R.~L. Schalek, J.~A. Conchello,
  S.~Knowles-Barley, D.~Lee, A.~Vázquez-Reina, V.~Kaynig, T.~R. Jones,
  M.~Roberts, J.~L. Morgan, J.~C. Tapia, H.~S. Seung, W.~G. Roncal, J.~T.
  Vogelstein, R.~Burns, D.~L. Sussman, C.~E. Priebe, H.~Pfister, and J.~W.
  Lichtman.
\newblock Saturated reconstruction of a volume of neocortex.
\newblock {\em Cell}, 162:648--661, 2015.

\bibitem{andres2012}
B.~Andres, T.~Kroeger, K.~L. Briggman, W.~Denk, N.~Korogod, G.~Knott, U.~Kothe,
  and F.~A. Hamprecht.
\newblock Globally optimal closed-surface segmentation for connectomics.
\newblock {\em ECCV}, 2012.

\bibitem{funke2012}
J.~Funke, B.~Andres, F.~A. Hamprecht, A.~Cardona, and M.~Cook.
\newblock Efficient automatic 3d-reconstruction of branching neurons from em
  data.
\newblock {\em Proc. of the IEEE Computer Soc. Conf. on Computer Vision and
  Pattern Recognition (CVPR)}, pages 1004--1011, 2012.

\bibitem{kaynig2015}
V.~Kaynig, A.~Vazquez-Reina, S.~Knowles-Barley, M.~Roberts, T.~R. Jones,
  N.~Kasthuri, E.~Miller, J.~Lichtman, and H.~Pfister.
\newblock Large-scale automatic reconstruction of neuronal processes from
  electron microscopy images.
\newblock {\em Medical Image Analysis}, 11(1):77--88, 2015.

\bibitem{gala2013}
J.~Nunez-Iglesias, R.~Kennedy, J.~Shi T.~Parag, and D.B. Chklovskii.
\newblock Machine learning of hierarchical clustering to segment 2d and 3d
  images.
\newblock {\em PLoS ONE}, 8(8), 2013.

\bibitem{amelio2011}
A.~Vazquez-Reina, D.~Huang, M.~Gelbart, J.~Lichtman, E.~Miller, and H.~Pfister.
\newblock Segmentation fusion for connectomics.
\newblock {\em Proceedings of the IEEE International Conference on Computer
  Vision (ICCV)}, 2011.

\bibitem{ciresan2012}
D.~C. Ciresan, A.~Giusti, L.~M. Gambardella, and J.~Schmidhuber.
\newblock Deep neural networks segment neuronal membranes.
\newblock {\em Adv. in Neural Information Processing Systems (NIPS)}, pages
  2852–--2860, 2012.

\bibitem{tschopp2016}
F.~Tschopp, J.~N.~P. Martel, M.~Cook S.~C.~Turaga, and J.~Funke.
\newblock Efficient convolutional neural networks for pixelwise classification
  on heterogeneous hardware systems.
\newblock {\em 13th IEEE Int. Sym. on Biomedical Imaging (ISBI)}, pages
  1225--1228, 2016.

\bibitem{boykov2001}
Y.~Boykov, O.~Veksler, and R.~Zabih.
\newblock Fast approximate energy minimization via graph cuts.
\newblock {\em IEEE Trans. on PAMI}, 23(11):1222--1239, 2001.

\bibitem{funke2014}
J.~Funke, J.~N.~P. Martel, S.~Gerhard, B.~Andres, D.~C. Cireşan, A.~Giusti,
  L.~M. Gambardella, J.~Schmidhuber, H.~Pfister, A.~Cardona, and M.~Cook.
\newblock Candidate sampling for neuron reconstruction from anisotropic
  electron microscopy volumes.
\newblock {\em International Conference on Medical Image Computing and
  Computer-Assisted Intervention (MICCAI)}, pages 17--24, 2014.

\bibitem{jain2011}
V.~Jain, S.~Turaga, K.~Briggman, M.~Helmstaedter, W.~Denk, and S.~Seung.
\newblock Learning to agglomerate superpixel hierarchies.
\newblock {\em Adv. in Neural Information Processing Systems (NIPS)}, 2011.

\bibitem{smola2007}
A.~Smola, S.V.N Vishwanathan, and Q.V. Le.
\newblock Bundle methods for machine learning.
\newblock {\em Adv. in Neural Information Processing Systems (NIPS)}, 21, 2007.

\bibitem{teo2010}
C.H. Teo, S.V.N Vishwanathan, A.~Smola, and Q.V. Le.
\newblock Bundle methods for regularized risk minimization.
\newblock {\em Journal of Machine Learning Research}, 11:311--365, 2010.

\bibitem{rand1971}
W.~Rand.
\newblock Objective criteria for the evaluation of clustering methods.
\newblock {\em Journal of the American Statistical Association}, 1971.

\bibitem{meila2005}
Marina Meila.
\newblock Comparing clusterings - an information based distance.
\newblock {\em Journal of Multivariate Analysis}, 98:873--895, 2007.

\bibitem{funke2016}
J.~Funke, F.~Moreno-Noguer, A.~Cardona, and M.~Cook.
\newblock Ted: A tolerant edit distance for segmentation evaluation.
\newblock {\em arXiv:1503.02291}, 2016.

\bibitem{ilastik2011}
C.~Sommer, C.~Straehle, U.~Koethe, and F.~A. Hamprecht.
\newblock ilastik: Interactive learning and segmentation toolkit.
\newblock {\em Eighth IEEE Int. Sym. on Biomedical Imaging (ISBI)}, pages
  230--233, 2011.

\end{thebibliography}

\end{document}